%% file: main.tex
\documentclass[11pt]{article}

\usepackage{acl}

\usepackage{times}
\usepackage{latexsym}

\usepackage[T1]{fontenc}

\usepackage{microtype}

\usepackage{xcolor}
\newcommand{\new}[1]{#1}

\usepackage{inconsolata}

\usepackage{tipa}

\usepackage[inkscape={/usr/bin/inkscape -z -C}]{svg}

\usepackage{stfloats}

\usepackage{graphicx}

\usepackage{tabularx}
\usepackage{siunitx}
\usepackage{float}

\usepackage{amsfonts}
\usepackage{amsmath}

\usepackage{subcaption}

\usepackage{makecell}
\usepackage{multirow} 
\usepackage[disable,backgroundcolor=gray!20,textsize=footnotesize]{todonotes}

\title{Curated Datasets and Neural Models for Machine Translation of Informal Registers between Mayan and Spanish Vernaculars}

\author{Andrés Lou, Juan Antonio Pérez-Ortiz, \\ {\bf Felipe Sánchez-Martínez,} {\bf Víctor M. Sánchez-Cartagena} \\[1ex]
        Dep. de Llenguatges i Sistemes Informàtics, Universitat d'Alacant \\ 
        E-03690 Sant Vicent del Raspeig, Spain\\[1ex]
        \texttt{and\_lou@ua.es}, \texttt{\{japerez,fsanchez,vmsanchez\}@dlsi.ua.es}}

\begin{document}
\maketitle
\begin{abstract}
The Mayan languages comprise a language family with an ancient history, millions of speakers, and immense cultural value, that, nevertheless, remains severely underrepresented in terms of resources and global exposure. In this paper we develop, curate, and publicly release a \new{set of corpora} in several Mayan languages spoken in Guatemala and Southern Mexico, \new{which we call MayanV}. The datasets are parallel with Spanish, the dominant language of the region, and are taken from official native sources focused on representing informal, day-to-day, and non-domain-specific language. As such, and according to our dialectometric analysis, they differ in register from most other available resources. Additionally, we present neural machine translation models, trained on as many resources and Mayan languages as possible, and evaluated exclusively on our datasets. We observe lexical divergences between the dialects of \new{Spanish} in our resources \new{and the more widespread written standard of Spanish}, and that resources other than the ones we present do not seem to improve \new{translation} performance, indicating that many such resources may not accurately capture common, real-life language usage. The MayanV dataset is available at \url{https://github.com/transducens/mayanv}.
\end{abstract}

\section{Introduction}
\label{sec:introduction}

The Mayan language family is 
spoken in an area covering the modern states of Guatemala, Belize, and Southern Mexico \citep[p. 81]{law2014language}. It consists of around 30 languages grouped into five or six major sub-groups, depending on the source \citep{campbell1985mayan,law2014language}. Most subgroups and most speakers are found today in Guatemala, where between 40-60\% of the population are native speakers \citep{censo2018,england2003mayan}. Mayan languages are relatively healthy, 
but their presence online and on the global scene in general is almost non-existent. In effect, Mayan languages, despite the total number of speakers, are considered to be somewhat in decline: according to \citet{richards2003atlas}, only around half the population of ethnic Mayas are Mayan speakers, and the languages are associated in many social contexts to 
backwardness, ignorance and poverty \citep{england2003mayan}. 

\begin{figure}
  \centering
  \includegraphics[scale=.18]{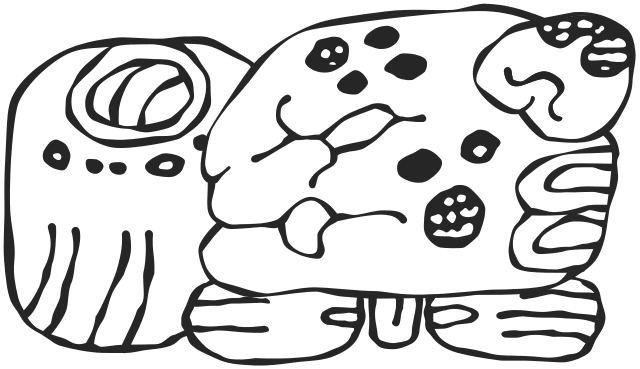}
  \caption{Sample of the ancient Mayan script, reading \textit{b'alam}, ``jaguar'', using a combination of the logogram and the syllabogram. Attribution: Goran tek-en under license \href{https://creativecommons.org/licenses/by-sa/4.0/deed.en}{CC BY-SA 4.0}.}
  \label{fig:balam}
\end{figure}

To begin addressing the problem of lack of representation 
in the digital realm, and
increase access to modern technology and information sources for indigenous communities, we seek to develop neural machine translation (NMT)~\citep{koehn_2020} systems. 
To
train and evaluate them, it is first necessary to produce and curate corpora of all the languages involved. In this paper, we present MayanV, a series of curated parallel corpora between various Mayan languages and Spanish. The language register in these datasets is informal, familial, and non-domain-specific, which best reflects the most 
common use of the languages involved. In general, online resources for building working NMT models are very scarce: the two greatest sources for parallel texts are the Bible, whose overly formal and potentially archaic language does not reflect most modern use cases, and the parallel texts of the Jehova's Witnesses website (jw.org), whose language, though much closer to modern usage, is still somewhat divorced from the common, day-to-day activities carried out by most Mayan speakers, as our dialectometric analysis suggests. Outside these two sources, bilingual and monolingual texts longer than a few thousand sentences are rare and not suited for processing, existing mostly as human-readable PDF files. Because of such scarcity of parallel resources for any  Mayan language, especially those with just a few thousand, or even a few hundred, speakers, we use the parallel corpora we have built to train and evaluate a number of bilingual and multilingual NMT systems; in particular, multilingual systems have proven effective when dealing with low-resource and underrepresented languages \citep{lakew2018comparison}. 


While other notable efforts in NMT of low-resource and endangered languages have been carried out recently, such as the No Language Left Behind (NLLB) \citep{team2022NoLL} and MADLAD projects \citep{Kudugunta2023MADLAD400AM}, these include little to no focus on Mayan languages. Our paper, in contrast, focuses on a more formal introduction of the Mayan languages to the larger \new{natural language processing} (NLP) community and on the presentation, curating, and release of parallel datasets that may be used for benchmarking future translation endeavours. \new{The corpora are available at \url{https://github.com/transducens/mayanv/}.}



The rest of the paper is organised as follows. The next section offers a historical and linguistic overview of the Mayan languages motivated by their relative obscurity amongst the NLP community. Section~\ref{sec:related_work} then presents the related work in the field. Section~\ref{sec:development_and_curation_of_the_MayanV} presents in detail the extraction and curating of resources for dataset creation and model training, \new{including a dialectometric analysis by which we characterise the dialectal and register divergence between the Spanish found in MayanV and the more standard variety of jw.org.} Afterwards, Section~\ref{sec:experimental_setup} describes and evaluates the NMT systems we have built. The paper ends with final remarks and a description of potential future research directions.

\section{Overview of the Mayan Languages} 
\label{sec:the_mayan_languages}

\begin{figure}[t]
  \centering
  \includesvg[scale=.7]{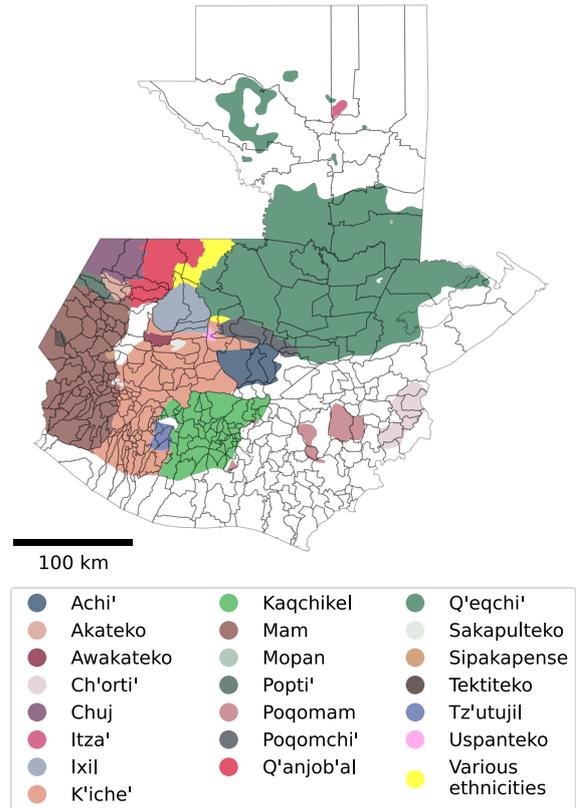}
  \caption{The Mayan linguistic communities of Guatemala}
  \label{fig:mayan_langs}
\end{figure}

The oldest attested Mayan language, referred to as Classic Maya, dates back to ca 300 BC. Written in the the Mayan script, \new{it belongs to a tradition corresponding to the} few instances in human history in which writing was independently invented, along with the systems developed in Ancient Egypt, Sumer, and Ancient China \citep[p 762]{fagan1996oxford}. Figure~\ref{fig:balam} shows an example of the script. The Mayan script is logosyllabic: glyphs may act either as logograms, representing a complete semantic unit, or syllabograms, representing a syllable \citep{coe2016reading}. \new{Modern Mayan languages are written using the Latin script, with additional diacritics used to denote features such as vowel length and glottalisation, amongst others.}

Much variation exists in terms of breadth and number of speakers. The three most spoken languages ---K'iche', Yucatec Mayan, and Q'eqchi--- all have between a million and half a million native speakers, and are spoken across different geographical areas in Guatemala and Mexico  \citep{richards2003atlas,law2014language}. In contrast, there are some languages with very few speakers and in imminent danger of language death, as is the case of Itza', which is only spoken by elderly adults and has fewer than 1\,000 native speakers \cite{censo2018,Eberhard2021ethnologue}. However, despite their size in terms of speakers and geographical extent, \new{as shown in Figure~\ref{fig:mayan_langs}}, Mayan languages are usually not given an official status in their respective countries; their use is widespread in daily activities and familial environments, but they are \new{nearly non-existent} 
in matters of governance, education, mass media, and healthcare \citep{romero2017labyrinth}. For example, in Guatemala, during 
Covid-19 pandemic, the official online portal to register for the government-sponsored vaccination program was only accessible in Spanish \citep{espana2021plan}; to this day it continues to lack any Mayan language version. Even when official support is said to exist, the actual implementation of educational \new{initiatives remain deficient}, as is the case with incipient government-sponsored programs to teach Yucatec Mayan in schools \citep{bote2023yucateco}. 

Bilingualism with Spanish is very common, though not uniform across geographical, population and gender lines: isolated communities often present a high number of monolingual speakers, and men usually exhibit higher proficiency in bilingualism \citep{Bennett2015Intro,romero2017labyrinth,richards2003atlas}. Language shift, whereby a Mayan language is displaced by either Spanish or, less commonly, another Mayan language \citep{Bennett2015Intro,romero2012they}, occurs more quickly in urban environments, where a lingua franca is expected to be used.

Literacy is overall low as a result of decades of policy where using native languages in schools was discouraged or even punished \citep{french2010maya}. As a result, many Mayan speakers regard the orthography of their own languages as more difficult and inaccessible than that of Spanish. Change was slowly brought about with the introduction of bilingual educational programs in the early 1990s, which finally led to a continued effort to revitalise the role of written Mayan languages. The Guatemalan Academy of Mayan Languages (ALMG), established in 1990, plays an important role in the standardisation of both the orthography of the different \emph{linguistic communities} and their corresponding spoken languages, while also engaging in literacy and publication efforts.\footnote{\url{https://www.almg.org.gt/nosotros/historia}} A similar role is played by the Proyecto Lingüístico Francisco Marroquín Foundation (PLFM),\footnote{\url{https://plfm.org/quienes-somos/historia}} also in Guatemala, and National Institute of Indigenous Languages (INALI), established in 2003, in Mexico,\footnote{\url{https://site.inali.gob.mx/Micrositios/normas/index.html}} both of which have published several grammars.

Mayan languages exhibit a high degree of dialectal variation \citep{romero2017labyrinth}. While the most important reason for this is natural language change and historical innovation, the sprachbund of Mayan languages in Guatemala and Southern Mexico has resulted in much linguistic exchange in the forms of loanwords and calques, usually manifesting as the influence of a language with more speakers and political leverage over another with fewer speakers and less influence. Dialectal divergence within the same language is also attested, as is the case with the dichotomies of the Western and Eastern, and Standard and Lowland dialects of Q'eqchi' \citep{dechicchis1989q,romero2012they}. Additionally, the politics of identity play a key role in delimiting the difference between a language and a dialect in the mind of their speakers, as seen in the cases of Achi, Akatek, and Chalchitek, which are sometimes considered dialects of K’iche’, Q’anjob’al, and Awakatek, respectively \citep{Bennett2015Intro}. In general, Mayan languages exhibit limited mutual intelligibility, and code-switching with Spanish and other Mayan languages in areas of high contact is common \citep{little2009language}.

Despite their relative obscurity in the field of NLP, Mayan languages are well studied and documented in matters of historical linguistics, morphosyntax, phonology, and semantics, as seen in extensive works such as those by \citet{Bennett2015Intro}, \citet{Bennet2015phonology}, \citet{coon2015morphosyntax}, \citet{henderson2015semantics}, \citet{polian2017morphology} and many others.

\section{Related Work} 
\label{sec:related_work}
African languages exist in a similar, albeit superlative, situation to that of Indigenous languages in the Americas. Thousands of African languages exist across the continent, boasting millions of speakers and serving as an important symbol of culture and cultural exchange; nevertheless, they are poorly represented in NLP applications. 
The Masakhane project \citep{Orife2020MasakhaneM} seeks to address this situation by fostering a community of researchers and non-researchers alike dedicated to advancing the development of NLP for African languages.
\citet{martinus2019focus} describes a number of challenges the African NLP community faces: Little official support for African indigenous languages; lack of resources for any kind of NLP task, and when those resources exist, they are hard to find; lack of benchmarks; and low reproducibility.

Mayan languages, and indigenous languages in general, face the same issues. Nevertheless, the work on NLP of Indigenous languages of the Americas continues \citep{mager2023proceedings}. Of note is the work by \citet{tyers2021corpus} and \citet{tyers2021survey}, who focus specifically on K'iche' and develop an annotated corpus for morphosyntactic structure and perform a survey of part-of-speech tagging methods respectively, and also \citet{pugh2023developing} who work on a finite-state transducer for performing morphological analysis on Yucatec Maya. Similarly to our work, \citet{oncevay-2021-peru} presents a multilingual NMT system, including both Spanish-to-many and many-to-Spanish models for Aymara, Ashaninka, Quechua, and Shipibo-Konibo, all indigenous languages spoken in Peru; in general, efforts to bring modern MT into the realm of endangered indigenous languages are gaining traction, with work focused on Nahuatl, Otomi, Guarani, Quechua, and many other prominent indigenous languages from countries where there exist a sizeable population of indigenous peoples \citep{mager-etal-2021-findings,parida-etal-2021-open,knowles-etal-2021-nrc,zheng-etal-2021-low,vazquez-etal-2021-helsinki}. Crucially, however, Mayan languages are \new{nearly non-existent} in these endeavours.

The introduction of multilingual and cross-lingual NMT \citep{10.1162/tacl_a_00065,conneau2019cross}, along with their application in low-resource scenarios \citep{lakew2018comparison,Karakanta2017NeuralMT,Madaan2020MultilingualNM} was of vital importance in the effort to improve NMT in languages with otherwise small or almost non-existent written bodies of work. Recent work on low-resource languages has been put forward by Meta's NLLB project \citep{team2022NoLL}, a contribution that includes several new benchmarks, including the \textsc{Flores+} dataset\footnote{\url{https://github.com/openlanguagedata/flores}} (\new{based on \textsc{Flores-200},} an update of \textsc{Flores-101} \citep{goyal2021flores}), and a state-of-the-art NMT translation model, called NLLB-200, focusing on underrepresented and low-resource languages, though, unfortunately, not a single Mayan language was included in their efforts. In contrast, Google's MADLAD-400 dataset, along with its accompanying translation model \citep{Kudugunta2023MADLAD400AM}, does include a number of Mayan languages in the form of monolingual corpora originating from jw.org and Bible sources.

\section{Development and Curation of MayanV} 
\label{sec:development_and_curation_of_the_MayanV}

\new{To develop the MayanV corpora}, we manually crawled, extracted, and cleaned \new{a number of} online resources, mostly published by the ALMG, except for the Tzeltal dictionary \citep{polian2018diccionario}; these extracted resources, which we collectively call the Mayan Vocabularies following the naming convention laid out by the ALMG, include the following languages: Achi \citep{acr_vocab}, Awakatek \citep{agu_vocab}, Chuj \citep{cac_vocab}, Itza' \citep{itz_vocab}, Ixil \citep{ixl_vocab}, Q'eqchi' \citep{kek_vocab}, Q'anjob'al \citep{kjb_vocab}, Mam \citep{mam_vocab}, Poqomam \citep{poc_vocab}, Poqomchi' \citep{poh_vocab}, K'iche' \citep{quc_vocab}, Sipakapense \citep{qum_vocab}, Tektitek \citep{ttc_vocab}, and Tz'utujil \citep{tzj_vocab}.

\subsection{Extracting the Mayan Vocabularies} 
\label{sub:extracting_the_MayanV}

\begin{figure} 
  \centering
  \begin{subfigure}[h]{.4\textwidth}
  \includegraphics[scale=.23]{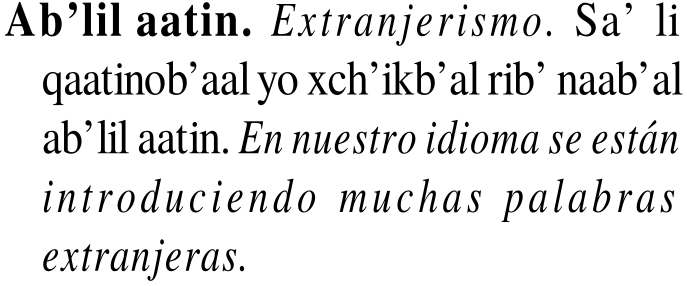}
    \caption{Original text, formatting, and layout.}
    \label{fig:kek_vocab}
  \end{subfigure}
  \begin{subfigure}[h]{.4\textwidth}
\vspace*{20pt}
    \begin{tabular}{l}
    \makecell[l]{{\small \texttt{Sa' li qaatinob'aal yo xch'ikb'al}} \\ {\small \texttt{rib'naab’al ab’lil aatin.}}} \\
    \makecell[l]{{\small \textit{En nuestro idioma se están introduciendo}} \\ {\small \textit{muchas palabras extranjeras}}} \\
    \end{tabular}
    \caption{Resulting parallel sentences.}
    \label{fig:kek_extracted}
  \end{subfigure}
  \caption{(\ref{fig:kek_vocab}) Entry in the Q'eqchi' corpus of MayanV. The Q'eqchi' term is in bold font; the first set of italics is the Spanish translation, ``loanword''; the regular text is a usage example of the term; the second set of italics is the Spanish translation of the example. (\ref{fig:kek_extracted}) Extracted Q'eqchi' sentence and its Spanish translation: ``There are many loanwords being introduced into our language.''}
  \label{fig:vocab_mam}
\end{figure}

The corpora in the Mayan Vocabularies can all be described as lists of entries, where each entry contains a Mayan word, its corresponding translation into Spanish, at least one example of the usage of the word in the Mayan language, and the corresponding translations into Spanish of such usages. Fourteen out of the fifteen corpora were only accessible as densely-formatted PDFs, the exception being the Tzeltal dictionary. See Figure~\ref{fig:vocab_mam} for an example of the format and Table~\ref{tab:vocabs} for a breakdown of the languages involved and the sizes of their respective corpus. \new{Using \texttt{pdfplumber},\footnote{\url{https://github.com/jsvine/pdfplumber}} the process by which we extracted the corpora can be summarised as follows: Using a heuristical approach involving the relative position of non-textual elements, such as margin delimiters or decorations, and that of typographically significant elements, e.g. the left-most bold-font character, each entry in the document was added to a list of bounding boxes and extracted. Words were then counted using white space, and sentences were split using punctuation and typography. Typically, each entry uses a particular font style for Mayan text and another different font style for Spanish text, as shown in Figure~\ref{fig:kek_vocab}. We were only interested in extracting full sentences, which were typically delimited by style and punctuation, and were often found in the latter parts of an entry. However, some entries included more than one example of usage, meaning than additional, oftentimes manual, inspection was required to extract the parallel text and avoid mixing up the languages.}

Using the Mayan Vocabularies, we developed a number of benchmarks intended to encourage other researchers to join the effort of developing NMT models for Mayan languages. \new{We have named the resulting dataset MayanV. The corpora that comprise MayanV} are part of the language and spelling standardisation efforts carried out by the ALMG. As such, in spite of the documented dialectal variation of some of the listed languages, we consider them good representations of modern and widespread language use.

\begin{table}[t]
\centering
\scalebox{.72}{
\begin{tabular}{l|crrr}
\hline
\textbf{ISO} & \textbf{Language} & \textbf{Words (Mayan)} & \textbf{Words (es)} & \textbf{Sentences} \\
\hline
acr & Achi        & 6\,994   & 7\,657   & 1\,343                         \\
agu & Awakatec   & 7\,325   & 9\,700   & 1\,930                          \\
cac & Chuj        & 9\,398   & 10\,916  & 2\,299                         \\
itz & Itza'       & 6\,069   & 7\,512   & 1\,539                         \\
ixl & Ixil        & 10\,888  & 12\,137  & 2\,325                         \\
kek & Q'eqchi'    & 18\,529  & 21\,835  & 4\,133                         \\
kjb & Q'anjob'al  & 18\,035  & 18\,238  & 3\,014                         \\
mam & Mam         & 15\,453  & 19\,117  & 3\,093                         \\
poc & Poqomam     & 18\,039  & 21\,744  & 3\,583                         \\
poh & Poqomchi'   & 6\,479   & 7\,149   & 1\,787                         \\
quc & K'iche'     & 14\,468  & 15\,474  & 2\,632                         \\
qum & Sipakapense & 9\,780   & 9\,328   & 1\,356                         \\
ttc & Tektitek   & 23\,571  & 24\,896  & 4\,022                         \\
tzh & Tzeltal     & 103\,309 & 128\,659 & 19\,846                        \\
tzj & Tz'utujil   & 12\,283  & 11\,404  & 2\,519                         \\
\hline
\end{tabular}
}
\caption{The 15 corpora of MayanV curated for our work. Size in terms of parallel sentences and words of the parallel corpora extracted from them. Sources for each corpus are discussed in the main text.
}
\label{tab:vocabs}
\end{table}

All mined PDFs are freely available for download at the ALMG's website, though there is considerable variation amongst them in terms of length, layout, typography, and content. 
This not only makes the extraction task laborious but also means that copy-editing and encoding errors are common in the original documents. \new{Indeed, post-extraction editing was necessary in several instances}. \new{For example, in the Mam (mam) corpus, the phonemic consonant /\textesh/ is represented with the wrong character (``õ'' instead of the digraph ``xh'') throughout the whole text, a fact that only came to light after thorough cross-examination with other corpora.} 


\subsection{Spanish Dialectometry of MayanV} 
\label{sec:dialectometric_analysis_of_the_MayanV}

\begin{table*}[t]
  \centering
  \scalebox{.73}{
  \input{figs-tabs/lexical_distances}
}
\caption{Average cosine distance and number of overlapping concepts between each of the Spanish texts in MayanV, including the dialect from jw.org when available. Columns n$_\text{jw}$ and jw denote the number of overlapping concepts and the average lexical distance between the Spanish in the corresponding MayanV corpus and the Spanish from jw.org. The upper diagonal of the table indicates the number of overlapping concepts between the Spanish of the respective MayanV corpora, while the lower diagonal indicates their lexical distance \new{(average cosine distance)}. Lower values indicate dialectal proximity.}
\label{tab:cosine_distance}
\end{table*}

The most salient trait of MayanV, standing in contrast to many of the available resources for MT, is the dialect and register in which both source and target languages are written. Much of the language in these texts is informal, day-to-day, and non-specific to any domain. Additionally, we note that much of the Spanish in the parallel texts is vernacular to Guatemala and Southern Mexico. As already mentioned, this is one of the most important aspects of MayanV as training and testing resources, as they reflect the most common use case amongst the marginalised indigenous minorities of Guatemala.

Following the example of \citet{donoso2017dialectometry}, we use the relative frequencies of the synonyms of a set of curated concepts taken from the Varilex project \citep{ueda2003varilex} to compute metrics that yield a sense of the lexical variation between contrasting dialects. As an example, the concept ``earthquake'' is materialised in the Spanish words \emph{movimiento}, \emph{movimiento sísmico}, \emph{remezón}, \emph{sacudida},  \emph{seísmo}, \emph{temblor}, \emph{temblor de tierra}, and  \emph{terremoto}. \new{We seek to compare the dialects of Spanish from MayanV and the Spanish used in jw.org. }

We represent each concept as a vector of the frequencies of each synonym. We then compute the average cosine distances amongst the Spanish texts of MayanV, and, when possible, the distance between the Spanish in MayanV and the jw.org corpora. For any two synonym vectors $\mathbf{s}_i, \mathbf{s}_j$ we compute the cosine distance as 

\[
1 - \frac{\mathbf{s}_i \cdot \mathbf{s}_j }{\lVert \mathbf{s}_i \rVert \lVert \mathbf{s}_j \rVert}.
\]

We only include vectors with at least one non-zero component, \new{and only compare concepts appearing in both corpora}. Results are shown in Table~\ref{tab:cosine_distance}. Lower values indicate dialectal proximity, though some \new{figures might be less indicative than others given the low number of overlapping concepts for any given pair of languages in general. Nevertheless, we observe evidence of considerable variation in the distances between these, possibly reflecting the lack of cross-linguistic regulation during the production of the documents from which MayanV was developed}. We also observe a noticeable divergence with the dialect of jw.org, which is empirically closer to a more widespread written standard.

The characterisation of the Spanish dialects involved in our task is of particular importance, given that the language acts as the most widespread common tongue in the region. The divergence of Spanish into several regional and mutually intelligible dialects is well-attested, even though the Spanish dialect of Guatemala remains somewhat understudied. Nonetheless, there have been important efforts towards its documentation~\citep{pato2023principales,Kotenyatkina2019LexicalPO}. We leverage the expertise of a Guatemalan team member, who, supported by the cited work, asserts that the Spanish dialectal variation in MayanV extends beyond lexical differences and accurately represents the prevalent \new{dialect} among rural populations in the country.


\section{Evaluating NMT Systems with MayanV}
\label{sec:experimental_setup}

\input{figs-tabs/corpora_table}
\input{figs-tabs/total_corpora}

\new{We develop bilingual and multilingual bidirectional NMT systems for Mayan languages and Spanish. Our aim is to assess the impact of the MayanV dataset on the translation quality of informal, familial domain texts. Thus, we consider baseline models that have not been exposed to this data during training and compare their performance to models trained with MayanV data. The validation and test sets consist exclusively of sentences from MayanV. Therefore, we will only evaluate language pairs for which MayanV data is available. Other languages still contribute to the training of multilingual models but are not explicitly evaluated. All corpora involved are parallel between Spanish and at least one Mayan language.}


\subsection{Experimental Settings}


\new{Bilingual and multilingual baseline models are trained using all relevant parallel corpora from OPUS, including Mozilla-I10n~\cite{tiedemann2004opus} and bible-uedin~\cite{christodouloupoulos2015massively}, as well as our own crawl of the Mayan versions of jw.org. Table~\ref{tab:combined_corpora} displays the data used to train the baseline models. To evaluate the impact of the MayanV dataset, we incorporate its data into the training pool (excluding dev and test sets), resulting in the setup shown in Table~\ref{tab:total_corpora}. We then train a new set of bilingual and multilingual models and compare them against the baseline models. Notice that, for some languages ---e.g. Itza' (itz)---, we are unable to produce baseline models since the only data available are those taken from MayanV.}

We use the same dev and test sets to, respectively, train and evaluate all models. We select 1\,000 sentences from each corpus in MayanV as individual test sets, and 1\,000 non-overlapping entries as dev sets;\footnote{All dev sets are combined into a single dev set for multilingual models.} in cases where there are less than 2\,000 entries ---e.g. Achi (acr) or Sipakapense (qum)--- we prioritise the test set and put the remaining sentences in dev set. \new{After this, all remaining instances, if any, are included in the train set of the non-baseline models.} Note that, as a result of the size of some corpora in MayanV, 
\new{not all training sets have a MayanV corpus associated to them.}



\new{In addition to models trained from scratch, we examine (baseline and non-baseline) bilingual models resulting from the fine-tuning of the \texttt{nllb-200-distilled-600M} model implementation from the Huggingface Transformers library~\cite{wolf2020transformers}. For this purpose, language tokens are added for each Mayan language, and NLLB-200's embedding layer is conveniently resized to accommodate them.}

Following the methods described by \citet{conneau2019cross} to address data imbalance in multilingual models, we rebalance our combined training corpus by fixing the size of the largest language fraction by number of sentences, i.e. Yucatec Mayan, and upsampling with replacement all other fractions by computing
\[
\lambda_i = \frac{p^\alpha_i}{\sum_{j=1}^N p_j^\alpha},
\]
where $p_i$ is the number of sentences of the $i-$th language, $N$ is the total number of languages in the corpus, and $\lambda_i$ is the resulting number of sentences of the $i$-th language once increased. We empirically determine the optimal value of the exponent to be~$\alpha=0.7$.


We used the \texttt{fairseq} toolkit version 0.12 \citep{ott2019fairseq} to carry out our experiments, \new{except for fine-tuning NLLB-200, for which we used Hugginface}. We used byte-pair encoding \citep{sennrich2016neural} to tokenise our datasets into subword units, learning the joint vocabulary between Spanish and the combined multilingual corpora over 60\,000 iterations. We used the Transformer model in its \texttt{base} configuration \citep[Table 3]{vaswani2017attention},\footnote{Approximately 94M parameters.} with the added difference of using 8\,000 warm-up steps and tied encoder-decoder embeddings. The multilingual models were trained on four parallel GPUs using mini-batches of 4\,000 tokens, \new{while the bilingual models were trained on a single GPU using mini-batches of 4\,000 tokens as well}. Validation was carried out every 5\,000 updates, and the patience, based on the BLEU score on the dev set, was set to 20 validation cycles in order to ensure optimisation for all languages involved; \new{each language pair was tested using the best-performing checkpoint as determined by validation}. We applied label smoothing with a value of $0.1$. \new{The NLLB-200 model was fine-tuned on a single GPU, using mini-batches of size 8, and a max sequence length of 1\,024 tokens. Validation was carried out every 500 steps, and the patience was set to 10 validation cycles. We used Adam for all training runs, with $\beta_1=0.9$ and $\beta_2=0.999$.}


\subsection{Results and Discussion} 
\label{sec:results_and_discussion}
\begin{table*}[h]
\centering
    \scalebox{.77}{\input{figs-tabs/bleu_table}}
\caption{BLEU scores for the bilingual and multilingual Mayan--Spanish and Spanish--Mayan translation tasks over baselines and models trained from scratch with and without using MayanV.}
\label{tab:results}
\end{table*}

\begin{table}[t]
    \centering
    \scalebox{.77}{\input{figs-tabs/bleu_table_nllb}}
    \caption{BLEU scores of the bilingual fine-tuned NLLB-200 model serving as baseline.}
    \label{table:results_nllb}
\end{table}

\new{Table~\ref{tab:results} shows the BLEU~\citep{papineni-etal-2002-bleu} scores\footnote{Computed using \texttt{sacrebleu}~\cite{post-2018-call} with signatures \texttt{nrefs:1|case:mixed|eff:no|tok:13a|smooth:exp| version:2.3.1} and \texttt{nrefs:1|case:mixed|eff:yes|nc:6| nw:0|space:no|version:2.3.1}} of all models trained from scratch, comparing baselines with those trained with MayanV, for the multilingual and bilingual translation systems; Table~\ref{table:results_nllb} shows the results of the corresponding baselines computed by fine-tuning the \texttt{nllb-200-distilled-600M} model. chrF2 scores, demonstrating similar trends, are presented in Appendix~\ref{sec:app_chrf2}.}

\new{The baseline models, which were trained over the available corpora that did not include any resources from MayanV, perform worse across the board when compared to models whose training data included MayanV. Within the bilingual runs, the NLLB-200 model fine-tuned to each individual language pair outperforms models trained from-scratch in almost all instances, with the notable exceptions being Tzeltal (tzh) and Q'eqchi' (kek). This increased performance is limited, however, since, as previously stated, NLLB-200 was not trained over any member of the Mayan language family, which limits the effects of positive transfer and curtails any other enhancement beyond the leverage of a few dozen loanwords taken from Spanish.} Overall, while it is possible to trivially assert that these results reflect the similarity, or lack thereof, between the data in the train and dev sets and the data in the test set, we argue 
that these results also reflect the considerable discrepancy between (1) formal and archaic registers, such as the ones we expect to find in the Bible; (2) domain-specific content, such as the religious education and world news content of jw.org, and the more quotidian register found in \new{MayanV, which is closer to the language Mayan speakers use in their day-to-day activities.} \new{These results also align with the dialectal divergences described in Section~\ref{sec:dialectometric_analysis_of_the_MayanV}. The comparison between baseline results and those of models that include MayanV is also similar for multilingual models.}


\new{When comparing bilingual and multilingual models, the latter outperform the former in all but one instance: Ixil (ixl) to Spanish.} For any given language with corpora other than MayanV, the results suggest the sizes of these corpora do not impact the performance of the translation task as much as we might expect; instead, the net size of the respective MayanV corpus seems to play a much greater role. For example, consider Mam (mam) and Tzeltal (tzh), whose jw.org corpora is similar in size, and, moreover, the former includes the New Testament in its training data; despite this, Tzeltal greatly outperforms Mam, seemingly because of the greater size of its MayanV corpus. Consider also Q'eqchi' (kek), with a sizeable jw.org corpus, the entirety of the Bible, and the second largest MayanV corpus; it, however, still compares to other languages \new{with a smaller representation in the multilingual model and whose MayanV corpus is of similar size}. 

Since the data used in the bilingual models is a subset of the data used in the multilingual model, we conclude that the inclusion of other languages in the training run acts as favourable leverage in that we are able to benefit from the performance boost and positive transfer of multilingual models. In the case of Tzeltal, since it corresponds by far to the largest corpus in MayanV, we suggest that, by simply having more instances to observe, the model is able to generalise more broadly, even in the face of different registers. Overall, these results align with previous findings \citep{Arivazhagan2019MassivelyMN}.




\section{Conclusions} 
\label{sec:conclusion}
Very little work has been done with indigenous American languages regarding their informal, day-to-day use and interaction with the dominant languages of their surroundings, and our results reflect this unfortunate fact. \new{We have developed and publicly released a curated set of parallel datasets between several Mayan languages and Spanish, which we call MayanV, focusing on the fact that the dialect and the register of the corpora is informal and non-domain-specific, which reflects a more common use case for the majority of native speakers. We train baseline bilingual and multilingual NMT Mayan-Spanish models from scratch, and fine-tune the NLLB-200 model for the bilingual case, and compare these with models whose train set is identical plus the addition of MayanV; we evaluate these models on a separate subset of MayanV and observe considerable improvements with respect to the baseline. This, along with a dialectometric analysis of the corpora involved, suggests that the vast majority of the available resources do not reflect the day-to-day usage of Mayan languages by their native speakers, nor do they facilitate the training and development of MT systems that might be useful for the most common use cases of the language.} However, we do observe several instances of improvement in performance when comparing multilingual models with their bilingual counterparts, suggesting that this remains a valid pathway for developing and training ever-improving NMT models for Mayan languages.

Future work in the area of NMT of Mayan languages should focus on the mining and production of datasets that reflect a closer use case to that which is useful for rural and often times marginalised indigenous communities, who usually do not speak using overly formal or archaic language, nor have consistent access to the internet and its associated zeitgeist. Finally, interesting work is to be done \new{by performing multilingual fine-tuning of larger pre-trained translation models, such as NLLB-200 itself, or MADLAD-400, whose purpose is to work with low-resource and endangered languages and which, sadly, minimise, or even outright exclude, Mayan languages in their current form}.

\section*{Acknowledgments}

Work funded by the Spanish Ministry of Science and Innovation, the Spanish Research Agency (AEI/10.13039/501100011033) and the European Regional Development Fund ``A way to make Europe'' through project PID2021-127999NB-I00. 

\section*{Limitations}
This paper has certain limitations, the primary being the relatively small size of the parallel corpora discussed therein. Despite significant efforts to construct this valuable resource, its limited scale may impact the generalisation and performance of any NMT systems developed for Mayan languages. The constrained amount of data available can potentially result in less robust models with limited capabilities to handle linguistic and contextual variability. Additionally, there is a risk that these systems could produce imprecise translations, posing potential safety and health concerns for users who rely on them.
Despite these limitations, the obtained results provide a solid and valuable foundation for future research and efforts to expand resources available for Mayan languages.



\section*{Ethics Statement}
The curated corpora and NMT models detailed in this paper actively advocate for the inclusion and promotion of indigenous languages. This aligns with the United Nations' goals, particularly during the International Decade of Indigenous Languages (2022-2032), emphasizing the crucial importance of preserving, revitalizing, and promoting indigenous languages globally. Our efforts will contribute to the development of language technology tools for Mayan languages, supporting linguistic diversity, and promoting equitable representation of language communities.

\bibliography{bibliography}
\appendix
\section{chrF2 scores}

\begin{table*}[b]
\centering
    \scalebox{.77}{\input{figs-tabs/chrf_table}}
\caption{chrF2 scores for the bilingual and multilingual Mayan--Spanish and Spanish--Mayan translation tasks over baselines and models trained from scratch with and without using MayanV.}
\label{tab:results_chrf}
\end{table*}

\label{sec:app_chrf2}

This appendix presents chrF2 scores~\citep{popovic-2017-chrf} of our baseline models, including both from-scratch and NLLB-200 approaches, as well as bilingual and multilingual configurations. The results presented in Table~\ref{tab:results_chrf} directly correspond to the BLEU scores detailed in Table~\ref{tab:results}, while Table~\ref{tab:results_nllb_chrf} offers scores for the NLLB-200 models, aligning with the BLEU results in Table~\ref{table:results_nllb}. The trends observed with both metrics are very similar, indeed aligning closely across all comparisons.

\begin{table}[!h]
    \centering
    \scalebox{.77}{\input{figs-tabs/chrf_table_nllb}}
    \caption{chrF2 scores of the bilingual fine-tuned NLLB model serving as baseline, to be compared with those in Table~\ref{tab:results_chrf}.}
    \label{tab:results_nllb_chrf}
\end{table}


\end{document}

%% file: figs-tabs/lexical_distances.tex
\begin{tabular}{l|ll|lllllllllllllll}
 & n$_\text{jw}$ & jw & acr & agu & cac & itz & ixl & kek & kjb & mam & poc & poh & quc & qum & ttc & tzh & tzj \\
 \hline
acr & - & - & - & 17 & 17 & 15 & 16 & 18 & 15 & 18 & 17 & 14 & 17 & 16 & 16 & 21 & 18 \\
agu & - & - & 0.175 & - & 21 & 16 & 16 & 21 & 18 & 19 & 19 & 15 & 20 & 20 & 18 & 21 & 19 \\
cac & - & - & 0.320 & 0.284 & - & 21 & 20 & 30 & 20 & 22 & 24 & 14 & 23 & 22 & 25 & 27 & 23 \\
itz & - & - & 0.241 & 0.252 & 0.295 & - & 15 & 23 & 14 & 17 & 18 & 14 & 16 & 16 & 17 & 19 & 20 \\
ixl & - & - & 0.227 & 0.202 & 0.293 & 0.238 & - & 19 & 16 & 18 & 21 & 13 & 18 & 17 & 21 & 22 & 19 \\
kek & 28 & 0.428 & 0.281 & 0.248 & 0.185 & 0.215 & 0.258 & - & 21 & 24 & 22 & 15 & 23 & 23 & 24 & 27 & 25 \\
kjb & - & - & 0.193 & 0.150 & 0.259 & 0.251 & 0.172 & 0.221 & - & 20 & 20 & 14 & 18 & 17 & 16 & 21 & 17 \\
mam & 23 & 0.402 & 0.188 & 0.164 & 0.269 & 0.224 & 0.181 & 0.212 & 0.115 & - & 20 & 15 & 21 & 19 & 18 & 24 & 20 \\
poc & - & - & 0.272 & 0.219 & 0.301 & 0.287 & 0.203 & 0.262 & 0.205 & 0.210 & - & 15 & 22 & 18 & 23 & 28 & 21 \\
poh & 12 & 0.313 & 0.168 & 0.142 & 0.274 & 0.194 & 0.180 & 0.232 & 0.158 & 0.149 & 0.184 & - & 14 & 15 & 14 & 17 & 14 \\
quc & 24 & 0.325 & 0.265 & 0.245 & 0.267 & 0.256 & 0.227 & 0.205 & 0.235 & 0.188 & 0.241 & 0.196 & - & 20 & 23 & 24 & 21 \\
qum & - & - & 0.293 & 0.200 & 0.303 & 0.266 & 0.264 & 0.251 & 0.254 & 0.204 & 0.316 & 0.201 & 0.184 & - & 21 & 24 & 21 \\
ttc & - & - & 0.354 & 0.271 & 0.316 & 0.299 & 0.219 & 0.233 & 0.291 & 0.277 & 0.256 & 0.245 & 0.216 & 0.294 & - & 26 & 19 \\
tzh & 38 & 0.387 & 0.359 & 0.264 & 0.338 & 0.364 & 0.283 & 0.304 & 0.265 & 0.242 & 0.252 & 0.246 & 0.281 & 0.319 & 0.292 & - & 24 \\
tzj & - & - & 0.231 & 0.227 & 0.285 & 0.223 & 0.190 & 0.211 & 0.214 & 0.186 & 0.240 & 0.191 & 0.191 & 0.228 & 0.291 & 0.310 & - \\
\end{tabular}

%% file: figs-tabs/corpora_table.tex
\begin{table*}[t]
\scalebox{.7}{
\begin{tabular}{l|rrr|lll|lll}
\hline
\textbf{ISO} &
  \multicolumn{3}{l|}{\textbf{jw.org}} &
  \multicolumn{3}{l|}{\textbf{Mozilla I10-n}} &
  \multicolumn{3}{l}{\textbf{bible-uedin}} \\ \hline
 &
  \multicolumn{1}{l}{\textbf{Words (Mayan)}} &
  \multicolumn{1}{l}{\textbf{Words (es)}} &
  \multicolumn{1}{l|}{\textbf{Sentences}} &
  \textbf{Words (Mayan)} &
  \textbf{Words (es)} &
  \textbf{Sentences} &
  \textbf{Words (Mayan)} &
  \textbf{Words (es)} &
  \textbf{Sentences} \\
  \hline
cak &
  716\,500 &
  620\,312 &
  54\,047 &
  \multicolumn{1}{r}{417\,839} &
  \multicolumn{1}{r}{310\,792} &
  \multicolumn{1}{r|}{28\,950} &
  \multicolumn{1}{r}{361\,971} &
  \multicolumn{1}{r}{186\,898} &
  \multicolumn{1}{r}{7\,862} \\
ctu &
  1\,536\,343 &
  1\,283\,405 &
  110\,521 &
   &
   &
   &
   &
   &
   \\
ixl &
  \multicolumn{1}{l}{} &
  \multicolumn{1}{l}{} &
  \multicolumn{1}{l|}{} &
  \multicolumn{1}{r}{17005} &
  \multicolumn{1}{r}{10\,286} &
  \multicolumn{1}{r|}{1\,955} &
   &
   &
   \\
kek &
  1\,134\,403 &
  1\,082\,829 &
  86\,612 &
   &
   &
   &
  \multicolumn{1}{r}{1\,157\,800} &
  \multicolumn{1}{r}{811\,163} &
  \multicolumn{1}{r}{31\,110} \\
mam &
  1\,711\,960 &
  1\,523\,974 &
  124\,051 &
   &
   &
   &
  \multicolumn{1}{r}{265\,748} &
  185\,460 &
  \multicolumn{1}{r}{7\,799} \\
poh &
  169\,965 &
  146\,517 &
  11\,330 &
   &
   &
   &
   &
   &
   \\
quc &
  993\,085 &
  928\,234 &
  83\,393 &
  \multicolumn{1}{r}{8\,498} &
  \multicolumn{1}{r}{5\,965} &
  \multicolumn{1}{r|}{661} &
  \multicolumn{1}{r}{312\,453} &
  \multicolumn{1}{r}{187\,684} &
  \multicolumn{1}{r}{7\,895} \\
tzh &
  1\,715\,549 &
  1\,457\,685 &
  120\,430 &
   &
   &
   &
   &
   &
   \\
tzo &
  3\,238\,511 &
  2\,942\,428 &
  234\,599 &
   &
   &
   &
   &
   &
   \\
yua &
  3\,554\,344 &
  3\,452\,737 &
  263\,500 &
  \multicolumn{1}{r}{4\,361} &
  \multicolumn{1}{r}{2\,440} &
  \multicolumn{1}{r|}{306} &
   &
   & \\
  \hline
\end{tabular}
}
\caption{Word and sentence distribution from the jw.org, 
Mozilla I10-n, and bible-uedin corpora.}
\label{tab:combined_corpora}
\end{table*}

%% file: figs-tabs/total_corpora.tex
\begin{table*}[t]
\centering
\scalebox{.7}{
\begin{tabular}{l|c|rrrrrr}
\hline 
\textbf{ISO} &
  \multicolumn{1}{l|}{\textbf{Language}} &
  \multicolumn{1}{l}{\textbf{Words (mayan)}} &
  \multicolumn{1}{l}{\textbf{Words (es)}} &
  \multicolumn{1}{l}{\textbf{Sentences}} & 
  \multicolumn{1}{c}{\textbf{train}} &
  \multicolumn{1}{c}{\textbf{dev}} &
  \multicolumn{1}{c}{\textbf{test}} \\ \hline
acr & Achi          & 6\,994     & 7\,657     & 1\,343 & -- & 343 & 1\,000   \\
agu & Awakatec     & 7\,325     & 9\,700     & 1\,930& -- & 930 & 1\,000      \\
cac & Chuj          & 9\,398     & 10\,916    & 2\,299& 299 (299) & 1\,000 & 1\,000   \\
cak & Kaqchikel     & 1\,496\,310 & 1\,118\,002 & 90\,859& 90\,859 & -- & --  \\
ctu & Ch'ol         & 1\,536\,343 & 1\,283\,405 & 110\,521& 110\,521 & -- & -- \\
itz & Itza'         & 6\,069     & 7\,512     & 1\,539& -- & 539 & 1\,000   \\
ixl & Ixil          & 27\,893    & 22\,423    & 4\,280& 2\,280 (325) & 1\,000 & 1\,000   \\
kek & Q'eqchi'      & 2\,310\,937 & 1\,915\,942 & 121\,883& 119\,883 (2\,133) & 1\,000 & 1\,000 \\
kjb & Q'anjob'al    & 18\,035    & 18\,238    & 3\,014& 1\,014 (1\,014) & 1\,000 & 1\,000   \\
mam & Mam           & 1\,727\,413 & 1\,543\,091 & 134\,943& 132\,943 (1\,093) & 1\,000 & 1\,000 \\
poc & Poqomam       & 18\,039    & 21\,744    & 3\,583& 1\,583 (1\,583) & 1\,000 & 1\,000   \\
poh & Poqomchi'     & 176\,444   & 153\,666   & 13\,117& 11\,117 & 787 & 1\,000  \\
quc & K'iche'       & 1\,328\,504 & 1\,137\,357 & 94\,581& 92\,581 (632) & 1\,000 & 1\,000  \\
qum & Sipakapense   & 9\,780     & 9\,328     & 1\,356& -- & 356 & 1\,000   \\
ttc & Tektitek     & 23\,571    & 24\,896    & 4\,022& 2\,022 & 1\,000 & 1\,000   \\
tzh & Tzeltal       & 1\,818\,858 & 1\,586\,344 & 140\,276& 138\,276 (17\,000) & 1\,000 & 1\,000 \\
tzj & Tz'utujil     & 12\,283    & 11\,404    & 2\,519& 519 & 1\,000 & 1\,000   \\
tzo & Tzotzil       & 3\,238\,511 & 2\,942\,428 & 234\,599& 234\,599 & -- & -- \\
yua & Yucatec Mayan & 3\,558\,705 & 3\,455\,177 & 263\,806& 263\,806 & -- & -- \\
\hline
Total & & 17\,331\,412& 15\,279\,230 & 1\,230\,470 & 1\,339\,332 & 12\,947 & 15\,000 \\
\hline
\end{tabular}
}
\caption{Complete multilingual corpus. Parentheses indicate the amount of training sentences taken from \new{MayanV}.}
\label{tab:total_corpora}
\end{table*}

%% file: figs-tabs/bleu_table.tex
\begin{tabular}{S|cccS|cccS}
\cline{2-9}
\multicolumn{1}{l}{{ \textbf{}}} &
  \multicolumn{4}{S|}{\textbf{maya-es}} &
  \multicolumn{4}{c}{\textbf{es-maya}} \\ \cline{2-9} 
\multicolumn{1}{l}{} &
  \multicolumn{2}{S|}{\textbf{bilingual}} &
  \multicolumn{2}{S|}{\textbf{multilingual}} &
  \multicolumn{2}{S|}{\textbf{bilingual}} &
  \multicolumn{2}{c}{\textbf{multilingual}} \\ \cline{2-9} 
\multicolumn{1}{l}{} &
  \multicolumn{1}{S|}{\textbf{Baseline}} &
  \multicolumn{1}{S|}{\textbf{MayanV}} &
  \multicolumn{1}{S|}{\textbf{Baseline}} &
  \textbf{MayanV} &
  \multicolumn{1}{S|}{\textbf{Baseline}} &
  \multicolumn{1}{S|}{\textbf{MayanV}} &
  \multicolumn{1}{S|}{\textbf{Baseline}} &
  \textbf{MayanV} \\ \hline
\multicolumn{1}{S|}{\textbf{acr}} &
  \multicolumn{1}{c|}{-} &
  \multicolumn{1}{c|}{-} &
  \multicolumn{1}{S|}{0.7} &
  2.3 &
  \multicolumn{1}{c|}{-} &
  \multicolumn{1}{c|}{-} &
  \multicolumn{1}{S|}{0} &
  0.1 \\
\multicolumn{1}{S|}{\textbf{agu}} &
  \multicolumn{1}{c|}{-} &
  \multicolumn{1}{S|}{0.0} &
  \multicolumn{1}{S|}{0.3} &
  1 &
  \multicolumn{1}{c|}{-} &
  \multicolumn{1}{S|}{0.0} &
  \multicolumn{1}{S|}{0.1} &
  0.3 \\
\multicolumn{1}{S|}{\textbf{cac}} &
  \multicolumn{1}{c|}{-} &
  \multicolumn{1}{S|}{0.1} &
  \multicolumn{1}{S|}{0.1} &
  4.2 &
  \multicolumn{1}{c|}{-} &
  \multicolumn{1}{S|}{0.0} &
  \multicolumn{1}{S|}{0} &
  3 \\
\multicolumn{1}{S|}{\textbf{itz}} &
  \multicolumn{1}{c|}{-} &
  \multicolumn{1}{c|}{-} &
  \multicolumn{1}{S|}{0.6} &
  0.9 &
  \multicolumn{1}{c|}{-} &
  \multicolumn{1}{c|}{-} &
  \multicolumn{1}{S|}{0.1} &
  0.1 \\
\multicolumn{1}{S|}{\textbf{ixl}} &
  \multicolumn{1}{S|}{0.3} &
  \multicolumn{1}{S|}{8.4} &
  \multicolumn{1}{S|}{0.2} &
  5.8 &
  \multicolumn{1}{S|}{0.0} &
  \multicolumn{1}{S|}{1.3} &
  \multicolumn{1}{S|}{0.1} &
  4.3 \\
\multicolumn{1}{S|}{\textbf{kek}} &
  \multicolumn{1}{S|}{0.6} &
  \multicolumn{1}{S|}{7.2} &
  \multicolumn{1}{S|}{2.4} &
  12.4 &
  \multicolumn{1}{S|}{1.1} &
  \multicolumn{1}{S|}{8.0} &
  \multicolumn{1}{S|}{2.8} &
  18.8 \\
\multicolumn{1}{S|}{\textbf{kjb}} &
  \multicolumn{1}{c|}{-} &
  \multicolumn{1}{S|}{3.5} &
  \multicolumn{1}{S|}{0.2} &
  5.8 &
  \multicolumn{1}{c|}{-} &
  \multicolumn{1}{S|}{4.0} &
  \multicolumn{1}{S|}{0} &
  7.1 \\
\multicolumn{1}{S|}{\textbf{mam}} &
  \multicolumn{1}{S|}{0.9} &
  \multicolumn{1}{S|}{3.8} &
  \multicolumn{1}{S|}{1.1} &
  6.5 &
  \multicolumn{1}{S|}{0.2} &
  \multicolumn{1}{S|}{1.0} &
  \multicolumn{1}{S|}{0.4} &
  2.9 \\
\multicolumn{1}{S|}{\textbf{poc}} &
  \multicolumn{1}{c|}{-} &
  \multicolumn{1}{S|}{5.2} &
  \multicolumn{1}{S|}{0.3} &
  9.3 &
  \multicolumn{1}{c|}{-} &
  \multicolumn{1}{S|}{1.8} &
  \multicolumn{1}{S|}{0} &
  6.4 \\
\multicolumn{1}{S|}{\textbf{poh}} &
  \multicolumn{1}{c|}{-} &
  \multicolumn{1}{S|}{0.2} &
  \multicolumn{1}{S|}{1.3} &
  5.5 &
  \multicolumn{1}{c|}{-} &
  \multicolumn{1}{S|}{0.3} &
  \multicolumn{1}{S|}{2.7} &
  4.8 \\
\multicolumn{1}{S|}{\textbf{quc}} &
  \multicolumn{1}{S|}{2.0} &
  \multicolumn{1}{S|}{2.9} &
  \multicolumn{1}{S|}{3} &
  7.8 &
  \multicolumn{1}{S|}{2.1} &
  \multicolumn{1}{S|}{3.3} &
  \multicolumn{1}{S|}{5.7} &
  10.1 \\
\multicolumn{1}{S|}{\textbf{qum}} &
  \multicolumn{1}{c|}{-} &
  \multicolumn{1}{c|}{-} &
  \multicolumn{1}{S|}{0.6} &
  1.6 &
  \multicolumn{1}{c|}{-} &
  \multicolumn{1}{c|}{-} &
  \multicolumn{1}{S|}{0.1} &
  0.1 \\
\multicolumn{1}{S|}{\textbf{ttc}} &
  \multicolumn{1}{c|}{-} &
  \multicolumn{1}{S|}{6.9} &
  \multicolumn{1}{S|}{0.7} &
  10.5 &
  \multicolumn{1}{c|}{-} &
  \multicolumn{1}{S|}{7.1} &
  \multicolumn{1}{S|}{0} &
  12.2 \\
\multicolumn{1}{S|}{\textbf{tzh}} &
  \multicolumn{1}{S|}{1.8} &
  \multicolumn{1}{S|}{68.1} &
  \multicolumn{1}{S|}{1.8} &
  70 &
  \multicolumn{1}{S|}{1.0} &
  \multicolumn{1}{S|}{58.0} &
  \multicolumn{1}{S|}{2.3} &
  64.1 \\
\multicolumn{1}{S|}{\textbf{tzj}} &
  \multicolumn{1}{c|}{-} &
  \multicolumn{1}{S|}{0.0} &
  \multicolumn{1}{S|}{0.2} &
  4.3 &
  \multicolumn{1}{c|}{-} &
  \multicolumn{1}{S|}{0.1} &
  \multicolumn{1}{S|}{0.1} &
  3.2 \\ \hline
\end{tabular}

%% file: figs-tabs/bleu_table_nllb.tex
\begin{tabular}{ccS|cc}
\cline{2-5}
                                  & \multicolumn{2}{S|}{\textbf{maya-es}}                    & \multicolumn{2}{c}{\textbf{es-maya}}                              \\ \cline{2-5} 
                                  & \multicolumn{1}{S|}{\textbf{Baseline}} & \textbf{MayanV} & \multicolumn{1}{S|}{\textbf{Baseline}} & \textbf{MayanV}          \\ \hline
\multicolumn{1}{S|}{\textbf{ixl}} & \multicolumn{1}{S|}{0.0}               & 1.2             & \multicolumn{1}{S|}{0.0}               & \multicolumn{1}{S}{4.6} \\
\multicolumn{1}{S|}{\textbf{kek}} & \multicolumn{1}{S|}{1.1}               & 9.7             & \multicolumn{1}{S|}{2.7}               & \multicolumn{1}{S}{5.4} \\
\multicolumn{1}{S|}{\textbf{mam}} & \multicolumn{1}{S|}{0.6}               & 5.2             & \multicolumn{1}{S|}{0.5}               & \multicolumn{1}{S}{0.9} \\
\multicolumn{1}{S|}{\textbf{quc}} & \multicolumn{1}{S|}{5.2}               & 5.7             & \multicolumn{1}{S|}{4.3}               & \multicolumn{1}{S}{5.4} \\
\multicolumn{1}{S|}{\textbf{tzh}} & \multicolumn{1}{S|}{3.5} & 51.1 & \multicolumn{1}{S|}{2.4} & \multicolumn{1}{S}{18.4} \\ \hline
\end{tabular}

%% file: figs-tabs/chrf_table.tex
\begin{tabular}{S|cccS|cccS}
\cline{2-9}
\multicolumn{1}{l}{{ \textbf{}}} &
  \multicolumn{4}{S|}{\textbf{maya-es}} &
  \multicolumn{4}{c}{\textbf{es-maya}} \\ \cline{2-9} 
\multicolumn{1}{l}{} &
  \multicolumn{2}{S|}{\textbf{bilingual}} &
  \multicolumn{2}{S|}{\textbf{multilingual}} &
  \multicolumn{2}{S|}{\textbf{bilingual}} &
  \multicolumn{2}{c}{\textbf{multilingual}} \\ \cline{2-9} 
\multicolumn{1}{l}{} &
  \multicolumn{1}{S|}{\textbf{Baseline}} &
  \multicolumn{1}{S|}{\textbf{MayanV}} &
  \multicolumn{1}{S|}{\textbf{Baseline}} &
  \textbf{MayanV} &
  \multicolumn{1}{S|}{\textbf{Baseline}} &
  \multicolumn{1}{S|}{\textbf{MayanV}} &
  \multicolumn{1}{S|}{\textbf{Baseline}} &
  \textbf{MayanV} \\ \hline
\multicolumn{1}{S|}{\textbf{acr}} &
  \multicolumn{1}{c|}{-} &
  \multicolumn{1}{c|}{-} &
  \multicolumn{1}{S|}{14.3} &
  18.4 &
  \multicolumn{1}{c|}{-} &
  \multicolumn{1}{c|}{-} &
  \multicolumn{1}{S|}{10.7} &
  12 \\
\multicolumn{1}{S|}{\textbf{agu}} &
  \multicolumn{1}{c|}{-} &
  \multicolumn{1}{S|}{2.5} &
  \multicolumn{1}{S|}{12.1} &
  14.8 &
  \multicolumn{1}{c|}{-} &
  \multicolumn{1}{S|}{1.1} &
  \multicolumn{1}{S|}{9.7} &
  13.1 \\
\multicolumn{1}{S|}{\textbf{cac}} &
  \multicolumn{1}{c|}{-} &
  \multicolumn{1}{S|}{2.1} &
  \multicolumn{1}{S|}{10.6} &
  21 &
  \multicolumn{1}{c|}{-} &
  \multicolumn{1}{S|}{1.9} &
  \multicolumn{1}{S|}{10} &
  27.2 \\
\multicolumn{1}{S|}{\textbf{itz}} &
  \multicolumn{1}{c|}{-} &
  \multicolumn{1}{c|}{-} &
  \multicolumn{1}{S|}{13.2} &
  14.8 &
  \multicolumn{1}{c|}{-} &
  \multicolumn{1}{c|}{-} &
  \multicolumn{1}{S|}{12.4} &
  13 \\
\multicolumn{1}{S|}{\textbf{ixl}} &
  \multicolumn{1}{S|}{2.5} &
  \multicolumn{1}{S|}{23.0 } &
  \multicolumn{1}{S|}{11.2} &
  22.8 &
  \multicolumn{1}{S|}{3.6} &
  \multicolumn{1}{S|}{10.5} &
  \multicolumn{1}{S|}{8.7} &
  20.7 \\
\multicolumn{1}{S|}{\textbf{kek}} &
  \multicolumn{1}{S|}{18.3 } &
  \multicolumn{1}{S|}{27.4 } &
  \multicolumn{1}{S|}{18.1} &
  32.6 &
  \multicolumn{1}{S|}{19,9 } &
  \multicolumn{1}{S|}{30.5 } &
  \multicolumn{1}{S|}{28.2} &
  42.9 \\
\multicolumn{1}{S|}{\textbf{kjb}} &
  \multicolumn{1}{c|}{-} &
  \multicolumn{1}{S|}{22.0} &
  \multicolumn{1}{S|}{13} &
  25.3 &
  \multicolumn{1}{c|}{-} &
  \multicolumn{1}{S|}{21.9} &
  \multicolumn{1}{S|}{11.1} &
  29.3 \\
\multicolumn{1}{S|}{\textbf{mam}} &
  \multicolumn{1}{S|}{15.8 } &
  \multicolumn{1}{S|}{24.0 } &
  \multicolumn{1}{S|}{16.6} &
  28.1 &
  \multicolumn{1}{S|}{19.9 } &
  \multicolumn{1}{S|}{23.1 } &
  \multicolumn{1}{S|}{22.3} &
  30.4 \\
\multicolumn{1}{S|}{\textbf{poc}} &
  \multicolumn{1}{c|}{-} &
  \multicolumn{1}{S|}{22.4} &
  \multicolumn{1}{S|}{13.2} &
  28.5 &
  \multicolumn{1}{c|}{-} &
  \multicolumn{1}{S|}{16.7} &
  \multicolumn{1}{S|}{10.4} &
  32 \\
\multicolumn{1}{S|}{\textbf{poh}} &
  \multicolumn{1}{c|}{-} &
  \multicolumn{1}{S|}{12.5} &
  \multicolumn{1}{S|}{15.8} &
  24.4 &
  \multicolumn{1}{c|}{-} &
  \multicolumn{1}{S|}{14.7} &
  \multicolumn{1}{S|}{24.2} &
  28.1 \\
\multicolumn{1}{S|}{\textbf{quc}} &
  \multicolumn{1}{S|}{19.9 } &
  \multicolumn{1}{S|}{20.8 } &
  \multicolumn{1}{S|}{20.7} &
  27.3 &
  \multicolumn{1}{S|}{24.2 } &
  \multicolumn{1}{S|}{25.1 } &
  \multicolumn{1}{S|}{30.9} &
  35.7 \\
\multicolumn{1}{S|}{\textbf{qum}} &
  \multicolumn{1}{c|}{-} &
  \multicolumn{1}{c|}{-} &
  \multicolumn{1}{S|}{15.6} &
  17.2 &
  \multicolumn{1}{c|}{-} &
  \multicolumn{1}{c|}{-} &
  \multicolumn{1}{S|}{10.2} &
  11.4 \\
\multicolumn{1}{S|}{\textbf{ttc}} &
  \multicolumn{1}{c|}{-} &
  \multicolumn{1}{S|}{26.8} &
  \multicolumn{1}{S|}{15} &
  31.4 &
  \multicolumn{1}{c|}{-} &
  \multicolumn{1}{S|}{27.3} &
  \multicolumn{1}{S|}{11} &
  34.7 \\
\multicolumn{1}{S|}{\textbf{tzh}} &
  \multicolumn{1}{S|}{15.3 } &
  \multicolumn{1}{S|}{75.2 } &
  \multicolumn{1}{S|}{16.7} &
  76.3 &
  \multicolumn{1}{S|}{22.4 } &
  \multicolumn{1}{S|}{71.6 } &
  \multicolumn{1}{S|}{25.6} &
  75.2 \\
\multicolumn{1}{S|}{\textbf{tzj}} &
  \multicolumn{1}{c|}{-} &
  \multicolumn{1}{S|}{3.1} &
  \multicolumn{1}{S|}{13.4} &
  21 &
  \multicolumn{1}{c|}{-} &
  \multicolumn{1}{S|}{5.2} &
  \multicolumn{1}{S|}{10.1} &
  23.9 \\ \hline
\end{tabular}

%% file: figs-tabs/chrf_table_nllb.tex
\begin{tabular}{ccS|cc}
\cline{2-5}
                                  & \multicolumn{2}{S|}{\textbf{maya-es}}                    & \multicolumn{2}{c}{\textbf{es-maya}}                               \\ \cline{2-5} 
                                  & \multicolumn{1}{S|}{\textbf{Baseline}} & \textbf{MayanV} & \multicolumn{1}{S|}{\textbf{Baseline}} & \textbf{MayanV}           \\ \hline
\multicolumn{1}{S|}{\textbf{ixl}} & \multicolumn{1}{S|}{9.3}               & 21.7            & \multicolumn{1}{S|}{5.0}               & \multicolumn{1}{S}{9.3}  \\
\multicolumn{1}{S|}{\textbf{kek}} & \multicolumn{1}{S|}{16.0}              & 33.7            & \multicolumn{1}{S|}{30.7}              & \multicolumn{1}{S}{37.8} \\
\multicolumn{1}{S|}{\textbf{mam}} & \multicolumn{1}{S|}{23.2}              & 27.6            & \multicolumn{1}{S|}{12.7}              & \multicolumn{1}{S}{24.7} \\
\multicolumn{1}{S|}{\textbf{quc}} & \multicolumn{1}{S|}{25.1}              & 26.5            & \multicolumn{1}{S|}{28.6}              & \multicolumn{1}{S}{32.4} \\
\multicolumn{1}{S|}{\textbf{tzh}} & \multicolumn{1}{S|}{20.8}              & 64.7            & \multicolumn{1}{S|}{28.2}              & \multicolumn{1}{S}{50.0} \\ \hline
\end{tabular}